\documentclass[10pt,twocolumn,letterpaper]{article}

\usepackage[pagenumbers]{iccv} 

\usepackage{pgfplots}
\usepackage{tikz}
\usetikzlibrary{calc}

\pgfplotsset{compat=1.18}

\definecolor{iccvblue}{rgb}{0.21,0.49,0.74}
\usepackage[pagebackref,breaklinks,colorlinks,allcolors=iccvblue]{hyperref}

\title{SA-LUT: Spatial Adaptive 4D Look-Up Table for Photorealistic Style Transfer}

\author{
Zerui Gong\textsuperscript{1}, Zhonghua Wu\thanks{Corresponding author: Z. Wu (e-mail: wuzhonghua@sensetime.com)}~~\textsuperscript{2}, Qingyi Tao\textsuperscript{2}, Qinyue Li\textsuperscript{2}, Chen Change Loy\textsuperscript{1} \\
\textsuperscript{1}S-Lab, Nanyang Technological University \\
\textsuperscript{2}SenseTime Research \\
{\tt\small \url{gong0060@e.ntu.edu.sg}
}}

\begin{document}
\maketitle

\begin{abstract}
Photorealistic style transfer (PST) enables real-world color grading by adapting reference image colors while preserving content structure.
Existing methods mainly follow either approaches: generation-based methods that prioritize stylistic fidelity at the cost of content integrity and efficiency, or global color transformation methods such as LUT, which preserve structure but lack local adaptability. To bridge this gap, we propose \textbf{Spatial Adaptive 4D Look-Up Table} (SA-LUT), combining LUT efficiency with neural network adaptability. SA-LUT features: (1) a Style-guided 4D LUT Generator that extracts multi-scale features from the style image to predict a 4D LUT, and (2) a Context Generator using content-style cross-attention to produce a context map. This context map enables spatially-adaptive adjustments, allowing our 4D LUT to apply precise color transformations while preserving structural integrity. 
To establish a rigorous evaluation framework for photorealistic style transfer, we introduce PST50, the first benchmark specifically designed for PST assessment. Experiments demonstrate that SA-LUT substantially outperforms state-of-the-art methods, achieving a 66.7\% reduction in LPIPS score compared to 3D LUT approaches, while maintaining real-time performance at 16 FPS for video stylization. Our code and benchmark are available at \url{https://github.com/Ry3nG/SA-LUT} 
\end{abstract}
\section{Introduction}
\label{sec:intro}

Photorealistic style transfer (PST) plays a vital role in film post-production and professional photography, requiring strict preservation of structural integrity while transferring color characteristics. Unlike artistic style transfer that tolerates distortions, PST demands faithful detail preservation and photorealistic fidelity \cite{luan2017deep}. Video applications further necessitate real-time processing and temporal consistency. These stringent requirements pose unique challenges for existing methods.

\begin{figure}[t!]
  \centering
  \includegraphics[width=0.95\columnwidth]{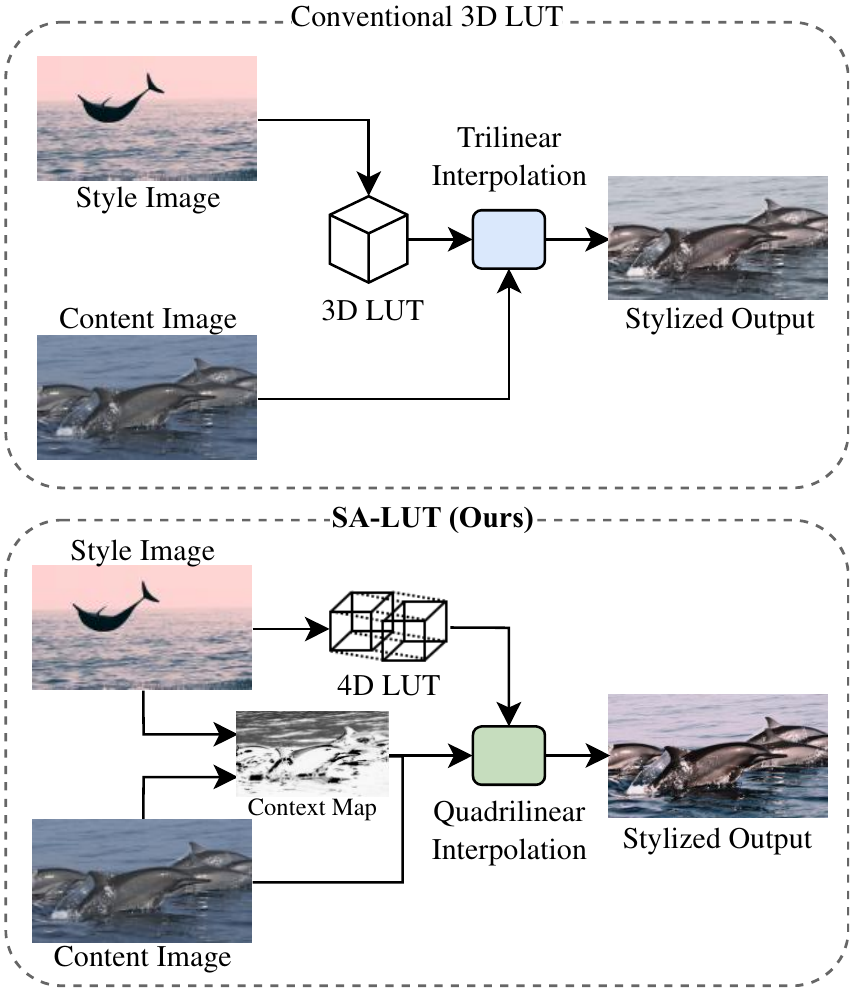}
  \caption{Comparison between conventional 3D LUT (uniform color mapping) and our SA-LUT (context-aware 4D LUT enabling spatially adaptive transformations).}
  \label{fig:teaser_pipeline}
\end{figure}

Existing methods exhibit critical trade-offs: Traditional color-transfer algorithms \cite{reinhard2001} induce inconsistent tinting through global statistics matching, while optimization-based techniques \cite{luan2017deep} impose prohibitive computational costs. While feed-forward networks \cite{WCT-NIPS-2017,yoo2019photorealistic} process images faster, they struggle with high-resolution inputs and heavy memory usage \cite{NeuralPreset}. Recent algorithms have improved efficiency \cite{Ho_2021_WACV, NeuralPreset} but typically lack the capacity for spatially adaptive adjustments.

Look-Up Tables (LUTs) offer an efficient alternative with hardware-friendly efficiency for professional workflows \cite{doi:10.2352/ISSN.2169-2629.2017.32.315,Karaimer2016}. While recent content-adaptive LUTs \cite{clutnet, Zeng2022, yang2022seplut, yang2022adaint, Li_2022_MuLUT} rival convolutional neural networks (CNNs) with reduced latency and memory usage \cite{yang2022adaint,yang2022seplut}, they only focus on single image enhancement~\cite{Zeng2022}. The specialized Neural 3D LUT (NLUT) \cite{chen2023nlut} pioneers style-aware mapping but inherits 3D LUT's fundamental constraint: identical transformation for same-color pixels across different semantic regions (\eg sky vs. sea). Recent advances \cite{gharbi2017deep,liu20234d} suggest incorporating contextual information into LUTs could be beneficial. However, these approaches have not been optimized for photorealistic style transfer's specific demands of maintaining structural integrity while enabling spatially adaptive color transformations.

We introduce Spatial Adaptive 4D Look-Up Table (SA-LUT), a framework for photorealistic style transfer optimized for professional workflows in log color space \cite{aces2017}. SA-LUT consists of two key components: (1) a \textit{Style-Weighted 4D LUT} that combines learned LUT bases using VGG-extracted style features, enabling style-guided color transformations, and (2) a \textit{Context Generator}, which produces a context map $\Gamma$ through content-style cross-attention. SA-LUT introduces a context-aware 4D LUT, concatenating the context map with the content image for quadrilinear interpolation. This enables precise region-specific color grading while maintaining structural integrity.

To establish standardized evaluation, we propose a PST50 Benchmark, the first comprehensive dataset with ground truth stylized images for objective assessment.
Experiments show SA-LUT outperforms existing state-of-the-art methods across all metrics. Our approach achieves a 66.7\% reduction in perceptual distance (LPIPS) between our stylized outputs and ground truth references compared to the previous LUT-based method \cite{chen2023nlut}, while also improving PSNR by 4.7dB and SSIM by 15\%. Moreover, our model can be generalized to video style transfer tasks. Specifically, our method can process 4K video at over 16 frames per second. This real-time performance, combined with our spatially adaptive capabilities, makes our method well-suited for professional applications requiring immediate visual feedback, such as on-set color grading or interactive content creation.
Our key contributions are:
\begin{itemize}
    \item To our knowledge, we are the first to propose a spatially adaptive 4D LUT model for photorealistic style transfer tasks. Our model can generate photorealistic images that accurately align with the given style images and preserve structural fidelity. Furthermore, our model also enables real-time inference on 4K footage at over 16 FPS.

    \item We introduce a Context Generator module to generate a context map that can be used to adaptively apply the 4D LUT. Specifically, we develop a cross-attention mechanism between content and style features, enabling distinct treatments for colors that appear the same but differ in semantic context or spatial positioning.

    \item To address the significant gap in photorealistic style transfer evaluation, we introduce PST50, the first benchmark with ground truth images and videos that enables both objective metric-based evaluation and perceptual assessment, establishing a new standard for the field.
\end{itemize}


\section{Related Work}
\label{sec:related_work}
\noindent\textbf{Photorealistic Style Transfer.}
Neural style transfer emerged with \citet{gatys2015neuralalgorithmartisticstyle, wu2019m2e, shi2021remember, zhong2023di, zhong2023sara, zhong2025ipvton, wu2025openuni}, defining style using Gram matrices of deep features from CNNs. While \citet{gatys2015neuralalgorithmartisticstyle}'s optimization-based approach was groundbreaking, its high computational cost spurred faster, feed-forward methods \cite{Johnson2016Perceptual,dumoulin2017learnedrepresentationartisticstyle}. However, these often produced painterly styles, less suited for photorealistic style transfer (PST).
\begin{figure*}[t]
  \centering
  \includegraphics[width=0.95\textwidth]{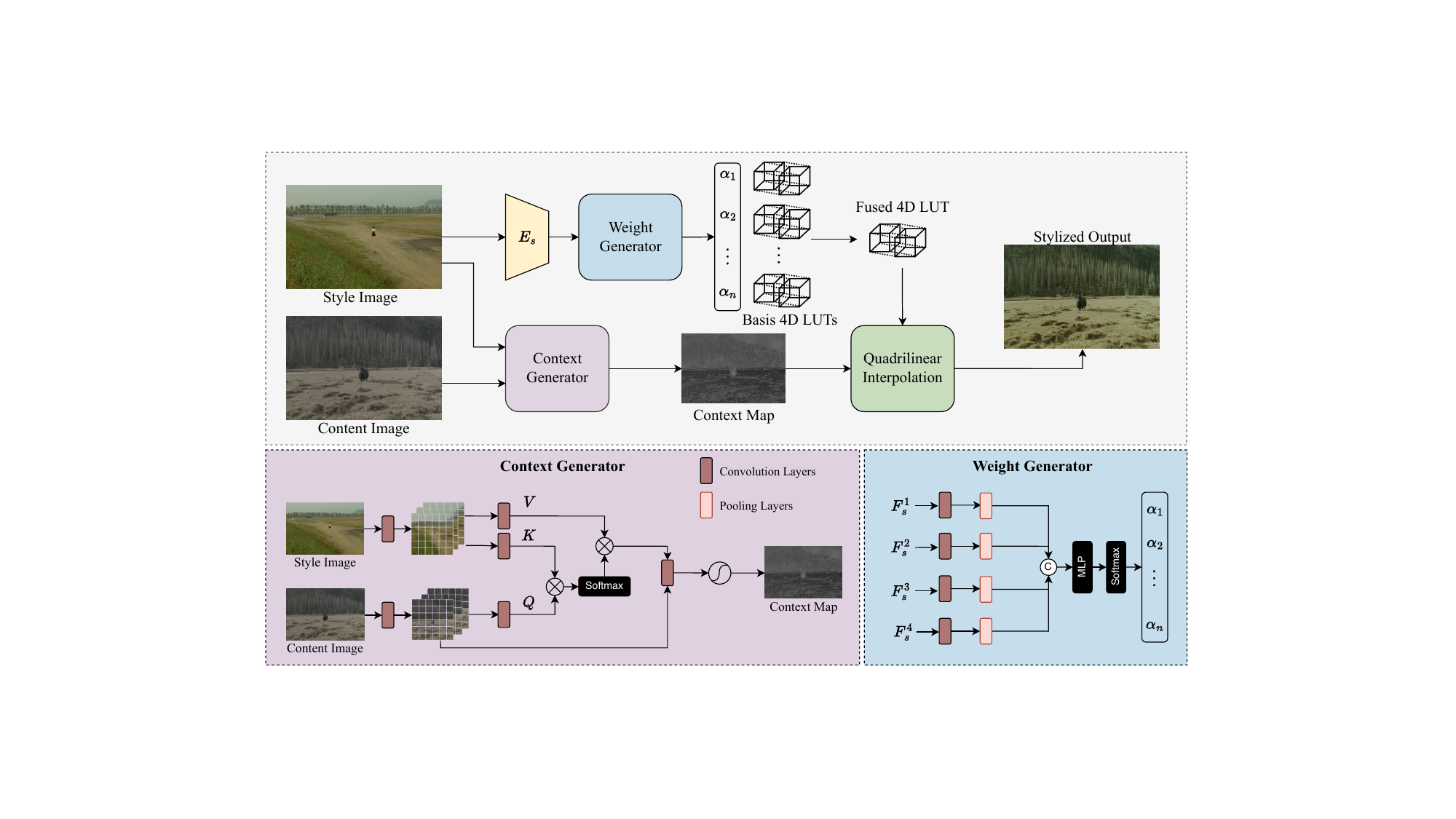}
  \caption{Overview of the SA-LUT framework. Our approach first constructs a style-guided 4D LUT by extracting style features from Style Encoder, $E_s$ (a pre-trained VGG network), refining these features, and predicting weights,  $\{\alpha_i\}^n_{i=1}$, for the 4D LUT bases. Simultaneously, the Context Generator produces a context map through cross-attention between content and style images. The final transformation applies the 4D LUT with spatial adaptivity using the context map and quadrilinear interpolation.}
  \label{fig:pipeline}
  \vskip -0.25cm
\end{figure*}

Dedicated PST methods aimed to bridge this gap. Deep Photo Style Transfer (DPST) by \citet{luan2017deep} used Matting Laplacian for spatial coherence, enhancing photorealism, but remained computationally intensive due to its optimization nature (minutes per image). Feed-forward methods improved efficiency: Whitening and Coloring Transform (WCT) \cite{WCT-NIPS-2017} achieved speed by aligning feature statistics, though its artistic style roots sometimes led to structural artifacts. PhotoWCT \cite{10.1007/978-3-030-01219-9_28} refined spatial details with unpooling and post-processing, and WCT$^2$ \cite{yoo2019photorealistic} employed wavelet operations for detail preservation, albeit with potential oversmoothing or memory demands at high resolutions.  These feed-forward methods directly predict pixel colors through deep networks, often incurring substantial computational overhead and potential distortions, especially when handling high-resolution images. More recently, diffusion-based models have shown promise in general image synthesis and some style transfer tasks. However, these methods are often orders of magnitude slower (e.g., \>5s per image for models like InstantStyle~\cite{wang2024instantstylefreelunchstylepreserving} or StyleID~\cite{chung2024styleinjectiondiffusiontrainingfree}) and can significantly alter content structure, placing them outside the specific goals of SA-LUT, which prioritizes real-time performance and strict structural preservation for photorealistic style transfer

\noindent\textbf{Preset-based and LUT-Based Image Processing.} Another line of work leverages image processing presets and Look-Up Tables (LUTs), widely adopted in professional color grading for their speed and hardware efficiency \cite{doi:10.2352/ISSN.2169-2629.2017.32.315,Karaimer2016}. Traditional 3D LUTs provide fast color transformations but are content-agnostic and apply uniform mappings \cite{Zeng2022}. Recent methods explore learnable presets or adaptive LUTs to incorporate content awareness, seeking a balance between performance and efficiency. Deep Preset \cite{Ho_2021_WACV} and Neural Preset \cite{NeuralPreset} learn global color adjustments, effectively acting as global presets to modify image colors.  Modulated Flow \cite{larchenko2024color} aligns color distributions globally using flow models. While offering efficiency, these global methods lack the flexibility to address spatially varying stylization needs.
To introduce content adaptivity into LUTs, \citet{Zeng2022} proposed image-adaptive 3D LUTs by blending multiple basis LUTs using a predictor network.  \citet{yang2022seplut} further enhanced LUT expressiveness by factorizing LUT operations into 1D and 3D components. In style transfer, Neural LUT (NLUT) \cite{chen2023nlut} fine-tunes a 3D LUT per video keyframe for efficiency, but its uniform mappings are spatially invariant, and test-time tuning limits rapid workflows. Several image processing techniques demonstrate the advantages of spatial adaptivity, including bilateral grids for HDR filtering \cite{gharbi2017deep}, cross-attention in AdaAttN for per-pixel style statistics \cite{liu2021adaattnrevisitattentionmechanism}, guided filtering for edge preservation \cite{6319316}, and higher-dimensional LUTs. Despite their benefits, these methods are not tailored to the specific needs of style-guided tasks. For instance, \citet{liu20234d} proposed a 4D LUT for image enhancement where both the LUT and its context dimension were derived solely from one image.
Our SA-LUT builds upon the concept of a 4D LUT but distinguishes itself by generating a style-conditioned 4D LUT and employing a context map derived from content-style cross-attention. This approach aims to merge the efficiency and photorealism of LUT-based methods with the spatial adaptivity necessary for precise, style-guided transformations, overcoming the limitations of both direct network-based PST methods and global preset/LUT approaches.

\noindent\textbf{Benchmark Dataset.}
PST evaluation lacks standardized benchmarks. Early works relied on small collections of image pairs without objective evaluation criteria~\cite{Ho_2021_WACV}. The MIT-Adobe FiveK dataset \cite{fivek}, while useful for image enhancement, represents general retouching rather than diverse stylistic changes. The Deep Photo Style Transfer (DPST) dataset \cite{luan2017deep} was an important step forward, providing photograph pairs for content and style images. However, it lacks ground truth stylized images, contains potential scene mismatches between content and style pairs, and occasionally incorporates artistic paintings, deviating from purely photorealistic transfer scenarios.
To address these limitations, we introduce PST50, providing both paired and unpaired evaluation protocols. Further details are discussed in Section~\ref{sec:pst50}.

\section{Method}
\label{sec:Methods}

We propose SA-LUT, a novel framework for photorealistic style transfer that leverages a spatially adaptive 4D Look-Up Table (LUT). Given a style images $I^{RGB}_{s}$, we first propose a Style-guided 4D LUT Generator to extract multi-scale features and predict a 4D LUT, $LUT_\text{fused}$, encoding style-specific color transformations. Then we design a Context Generator that computes a context map $\Gamma$ for the content image $I^{LOG}_{c}$ through cross-attention between content and style features, enabling region-aware color adjustments. Finally, our method applies $LUT_\text{fused}$ to the content image $I^{LOG}_{c}$ guided by its context values, achieving spatially-varying color transformations while preserving structural integrity. Figure~\ref{fig:pipeline} illustrates our complete pipeline. To effectively train our SA-LUT model, we design a specialized strategy that leverages synthetic and real-world style images with perceptual and adversarial losses.

\subsection{Style-Guided 4D LUT Generation}
\label{sec:style_weighted_4d_lut}

Our Style-Guided 4D LUT Generator, as shown in Figure~\ref{fig:pipeline} (top), creates a customized 4D LUT representing color transformation for each style image through two main steps: (1) a Style Encoder and Weight Generator that extracts style features and predicts style-specific weights, and (2) a LUT Fusion module that combines learnable basis LUTs according to these weights.

\subsubsection{Style Encoder and Weight Generator}
Following established approaches \cite{huang2017adain, chen2023nlut}, we extract multi-scale features from the style image $I^{RGB}_s$ using a VGG network. We obtain feature maps from four different layers of the VGG encoder (${F_{s}^{(1)}, F_{s}^{(2)}, F_{s}^{(3)}, F_{s}^{(4)}}$), capturing style information from low-level textures to high-level semantic content.

The Weight Generator processes these features through convolution layers and pooling operations to create a compact representation. The pooled features are concatenated and processed by an MLP with softmax activation:
\begin{equation}
  f_{\text{concat}} = \operatorname{Concat}_{d=1}^4\bigl(\operatorname{Pool_{max}}(\operatorname{Conv}(F_{s}^{(d)}))\bigr);
\end{equation}
\begin{equation}
\alpha = \operatorname{Softmax}\bigl(\mathrm{MLP}(f_{\text{concat}})\bigr).
\end{equation}
This generates a weight vector $\alpha\in\mathbb{R}^{N}$, where $N$ is the number of LUT bases in our model.

\subsubsection{LUT Fusion}
The LUT Fusion module combines multiple learnable basis LUTs to create a style-specific 4D LUT. Our model maintains a set of $N$ learnable 4D LUTs, each denoted as $\text{LUT}_i \in \mathbb{R}^{3 \times 2 \times D \times D \times D}$, where the dimensions correspond to RGB channels, context bins, and the discretization of the RGB color space (with resolution $D$). These basis LUTs are initialized randomly and optimized during training to represent diverse color transformation patterns. The choice of two context bins (representing two 3D LUT `slices') balances between expressive power and efficiency. Our continuous context map (Sec.~\ref{sec:context_gen}) and subsequent quadrilinear interpolation (Sec.~\ref{sec:quadrilinear}) effectively create a dense continuum of transformations by blending these two anchor LUT volumes.

Then, using the weight vector $\alpha$ estimated by the Weight Generator, we compute the style-specific 4D LUT as a weighted combination:
\begin{equation}
LUT_{\text{fused}} = LUT_{\text{identity}} + \sum_{i=1}^{N} \alpha_i \cdot LUT_i,
\label{eq:lut_fusion}
\end{equation}
where $LUT_{\text{identity}}$ serves as a residual connection ensuring that when $\alpha$ approaches zero, the transformation preserves the original input. The resulting fused LUT is clamped to $[0,1]$ to ensure valid color values.

This approach enables our model to generate a customized 4D LUT for each style image, effectively encoding its unique color transformation characteristics while maintaining flexibility through the learned basis LUTs.


\begin{figure*}[ht!]
  \centering
  \includegraphics[width=0.95\textwidth]{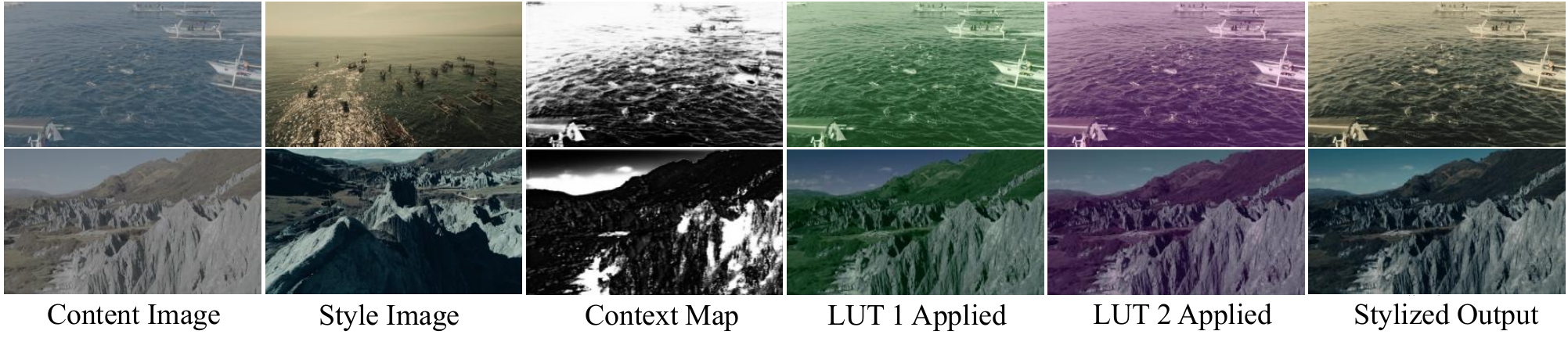}
  \caption{Visualizing intermediate results produced by our spatially adaptive 4D LUT approach. Our SA-LUT learns distinct color transformations with two 3D LUTs, balancing between expressive power and efficiency. Through context-guided interpolation between the two 3D LUT slices (Sec.~\ref{sec:context_gen}) and subsequent quadrilinear interpolation (Sec.~\ref{sec:quadrilinear}), our method achieves a more refined stylization that adapts to local image characteristics.}
  \label{fig:lut_visualization}
\end{figure*}

\subsection{Context Generator}
\label{sec:context_gen}
To enable spatially-varying stylization, we introduce a Context Generator that produces a content-specific context map. This component elevates our approach beyond global color transformations toward region-aware stylization that respects the semantic structure of both content and style images.

The Context Generator creates a context map $\Gamma$ that identifies corresponding regions between content and style images that should undergo similar color transformations. We first process both the content image $I^{LOG}_c$ and style image $I^{RGB}_s$ through lightweight CNN encoders composed of instance normalization and residual blocks.

Using these feature representations, we compute cross-attention between content and style features. Content features serve as queries ($Q$), while style features provide the keys and values ($K$ and $V$):

\begin{equation}
\text{Attn}(Q,K) = \operatorname{Softmax}\left(\frac{Q K^{\top}}{\sqrt{d}}\right).
\end{equation}
This attention mechanism establishes region-wise correspondences between content and style images. The resulting attended features are fused with the original content features through a convolutional module, followed by upsampling operations and a final convolution layer. This process generates a single-channel context map $\Gamma \in [0,1]^{H \times W}$ with the same spatial resolution as the content image, where each pixel value indicates the appropriate interpolation factor for that specific location.

\subsection{Quadrilinear Interpolation}
\label{sec:quadrilinear}
Unlike traditional 3D LUTs that apply uniform color transformations globally, our approach leverages a 4D LUT \cite{liu20234d} with an additional context dimension for spatially-varying transformations.

For each pixel, we apply a transformation that adaptively blends between two distinct 3D LUT color transformations based on its corresponding context value in $\Gamma$. This enables region-specific color adjustments while maintaining photorealistic appearance and smooth transitions between regions.

The spatially-adaptive application is implemented through quadrilinear interpolation:

\begin{equation}
I_p^{RGB} = \text{Quad}(LUT_{\text{fused}}, [\Gamma, I_c^{LOG}])
\end{equation}
Specifically, this process concatenates the context map $\Gamma$ with the content image $I_c^{LOG}$ to form a 4-channel input tensor, which is then processed through our fused 4D LUT. This approach computes a weighted average of the 16 nearest grid points in the 4D space for each pixel, ensuring smooth transitions across both spatial regions and color values.


Figure~\ref{fig:lut_visualization} illustrates SA-LUT's spatial adaptivity through the context maps. In the sea scene (top), the context map differentiates darker and lighter sea areas, resulting in distinct localized color grading. Similarly, in the mountain scene (bottom), it distinguishes darker sky/mountain regions from lighter ones. Hence, the context map intelligently identifies both semantic regions and luminance variations, enabling nuanced color transformations within the same object class, resulting in more refined and contextually accurate stylization.

\subsection{Training Strategy and Losses} \label{sec:training}
Photorealistic style transfer faces a fundamental challenge: the absence of standard datasets with ground truth examples. We address this through a three-part solution: (1) a dual-stream data approach combining synthetic examples with real-world images, (2) a patch similarity discriminator that effectively assesses style correspondence, and (3) a balanced combination of perceptual, regularization, and adversarial losses. This integrated approach enables SA-LUT to achieve both precision in color transformation and adaptability to diverse artistic styles.

\subsubsection{Data Preparation}
To overcome the absence of ground truth data, we develop a dual-stream training strategy:

\noindent\textbf{Synthetic Style Training.}
We create synthetic training data by applying professional 3D LUTs to LOG-space images. For each training instance, we select two LOG-space images ($I_c^{LOG}$ for content and $I_s^{LOG}$ for style base) and apply an identical 3D LUT to both, generating a ground truth image $I_c^{RGB}$ and a style reference $I_s^{RGB}$. During training, our network transforms $I_c^{LOG}$ using style guidance from $I_s^{RGB}$, with direct supervision against $I_c^{RGB}$. This approach provides precise color transformation supervision, though limited to the artistic range of available LUTs.

\noindent\textbf{Real Style Training.}
To incorporate authentic photographic aesthetics, we develop a Style2Log model that transforms RGB images into LOG space representations (details in supplementary materials). For each iteration, we first divide a single photograph into two non-overlapping crops $I^{RGB}_1$ and $I^{LOG}_2$. Then the $I^{RGB}_1$ becomes our style reference $I_s^{RGB}$, while $I^{RGB}_1$ is processed through Style2Log to create a synthetic content input $I_c^{LOG}$. Without ground truth for supervision, we employ adversarial training to guide stylization. This approach enables learning from diverse professional photography styles beyond predefined LUTs.


\subsubsection{Training Objectives and Discriminator}
We employ a specialized discriminator and targeted loss functions to guide the training process:

\noindent \textbf{Color Style Discriminator.}
To evaluate stylistic similarity without relying on pixel-level comparisons, we develop a multi-scale patch similarity discriminator. Unlike traditional GANs that classify images as real or fake, our discriminator quantifies style correspondence between generated outputs and reference images. The architecture employs three convolutional stages to extract hierarchical features, followed by computing correlation matrices between style and prediction features at each scale. The final similarity score aggregates these correlations, providing a robust measure of style transfer quality. For negative examples during training, we create mismatched style-prediction pairs by shuffling batch indices.

During the model training, our objective function combines three complementary components:

\noindent \textbf{Perceptual Loss.} For synthetic style training, we use LPIPS~\cite{zhang2018unreasonableeffectivenessdeepfeatures} to measure perceptual differences between predicted stylizations and ground truth images.

\noindent \textbf{Regularization Losses.} Following Zeng et al.~\cite{Zeng2022}, we apply total variation ($\mathcal{L}_{TV}$) and monotonicity ($\mathcal{L}_{MN}$) losses to ensure LUT smoothness and prevent color inversions.

\noindent \textbf{Adversarial Loss.} Our discriminator guides the model toward producing stylistically coherent results, particularly crucial for the real style stream where no ground truth exists.

Our final objective combines these components with balanced weighting:
\begin{align}
\mathcal{L}_\text{total} = \lambda_\text{1}\mathcal{L}_\text{lpips} + \lambda_\text{2}\mathcal{L}_\text{TV} + \lambda_\text{3}\mathcal{L}_\text{MN} + \lambda_\text{4}\mathcal{L}_\text{adv}.
\end{align}

\section{PST50 Benchmark Dataset}
\label{sec:pst50}

To address the lack of standardized evaluation in photorealistic style transfer (PST), we introduce PST50, a benchmark dataset for both paired and unpaired evaluation.

\begin{figure}[t]
  \centering
  \includegraphics[width=\columnwidth]{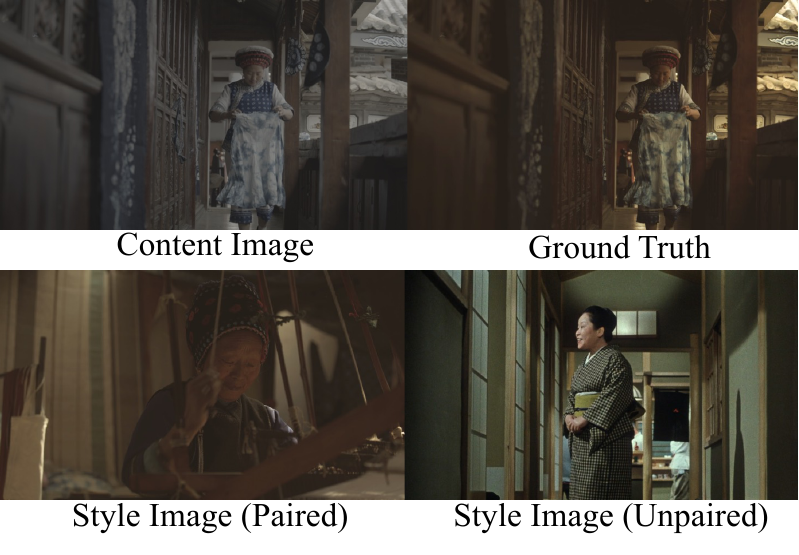}
  \caption{An example from our PST50 dataset.}
  \label{fig:pst50_visualize}
\end{figure}

\subsection{Dataset Collection and Scope}
PST50 contains 100 content-style image pairs (50 pair for each partition) from professional sources, featuring 50 diverse content images in four categories: natural landscapes (44\%), human/cultural subjects (26\%), architectural scenes (20\%), and wildlife photography (10\%). For paired evaluation, content and style images are from Mediastorm's professional footage library, ensuring high quality. Unpaired evaluation uses style references supplemented from expert color-graded cinema and documentary footage.

All 1080p images balance detail and processing, with approximately one-third featuring low-light scenes. Content videos are included for temporal consistency evaluation. This dual protocol (paired and unpaired) enables more comprehensive assessment than single-protocol datasets.

Compared with DPST \cite{luan2017deep}, the closest existing dataset, which contains 60 content-style pairs but lacks ground truth and includes non-photorealistic styles, limiting objective evaluation. PST50 prioritizes curated, high-quality professional imagery over potentially redundant samples, aligning with visual quality assessment practices \cite{SSIM}. To further quantify the diversity of PST50, we computed the average pairwise CIELAB Bhattacharyya distance for the color distributions of style images. As shown in Table~\ref{tab:color_diversity}, PST50 demonstrates greater inter-image color diversity compared to the DPST dataset across all color channels. This indicates that PST50 provides a challenging benchmark with a wider range of stylistic color variations.

\begin{table}[h]
\centering
\caption{Color diversity comparison using average pairwise CIELAB Bhattacharyya distance}
\label{tab:color_diversity}
\begin{tabular}{lccc}
\toprule
\textbf{Dataset} & \textbf{L* Dist} & \textbf{a* Dist} & \textbf{b* Dist} \\
\midrule
PST50 Paired & 1.759 & 2.676 & 2.892 \\
PST50 Unpaired & 1.491 & 2.499 & 2.811 \\
DPST \cite{luan2017deep} & 1.491 & 2.477 & 2.558 \\
\bottomrule
\end{tabular}
\end{table}

\subsection{Ground Truth Generation}

The ground truth images for the paired partition were created through a professional color grading workflow. We first applied initial color transformation using professional LUTs, then refined the color grading in DaVinci Resolve following industry practices. Finally, we performed manual adjustments for optimal color, contrast, and tone while maintaining photorealism. This process mirrors professional post-production, ensuring groundtruth represents achievable, high-quality results, balancing style fidelity and photorealism.

\section{Experiments}
\label{sec:experiments}

\begin{figure*}[t!]
  \centering
  \includegraphics[width=0.95\textwidth]{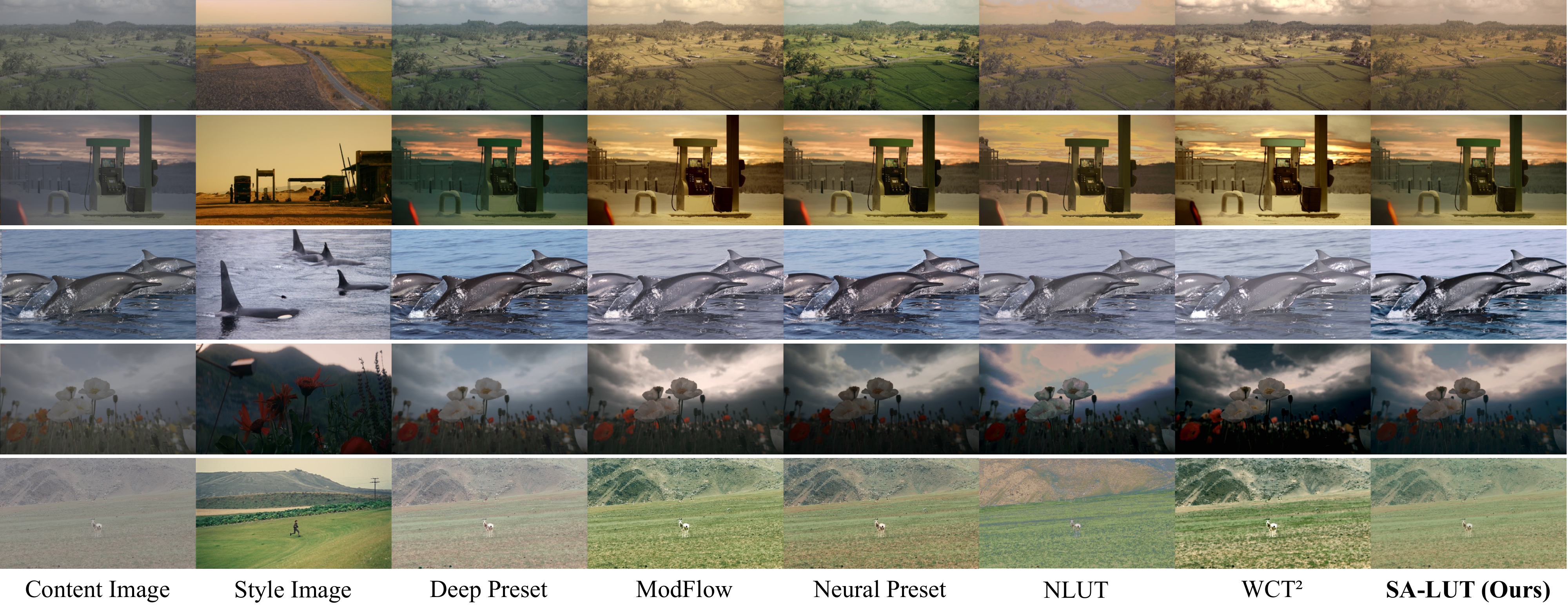}
  \caption{Visual comparison of photorealistic style transfer results with different methods.}
  \label{fig:visualize1}
\end{figure*}

\subsection{Implementation Details}

The model utilizes a 4D LUT with dimension parameters $D=17$ for RGB dimensions and 2 for the context dimension, with $K=64$ basis LUTs. We choose $D=17$ to achieve precise color mapping while maintaining computational efficiency. It's also a resolution consistent with standard color grading practices. During the model training, we use a warmup scheduling strategy for the generator, gradually increasing the learning rate during the initial training epochs, followed by cosine annealing for both the generator and the discriminator. We maintain training stability by using a lower learning rate for the discriminator (1/10 of the generator's rate). All experiments are performed on an NVIDIA RTX 3090 GPU.

\subsection{Comparisons with State-of-the-Art}

We compare SA-LUT against 5 state-of-the-art photorealistic style transfer methods: NLUT~\cite{chen2023nlut}, Neural Preset~\cite{NeuralPreset}, Deep Preset~\cite{Ho_2021_WACV}, WCT$^2$~\cite{yoo2019photorealistic}, and ModFlow~\cite{larchenko2024color}. We also include AdaIN~\cite{dumoulin2017learnedrepresentationartisticstyle} representing early work in neural style transfer that, while primarily designed for artistic stylization, serves as an important baseline to demonstrate advances in the field. We used official implementations for all methods and converted content images to rec.709 color space for fair comparison.

\subsubsection{Quantitative Evaluation}
\label{sec:quantitative}

We quantitatively evaluate performance using complementary metrics: LPIPS \cite{zhang2018unreasonableeffectivenessdeepfeatures} (perceptual loss to the groundtruth images $\downarrow$), PSNR (pixel fidelity $\uparrow$), SSIM \cite{SSIM} (structural preservation,$\uparrow$) and H-Corr ( \cite{gpass1996, Ho_2021_WACV} (color distribution similarity to style $\uparrow$). Moreover, we also conduct a user study for subjective perceptual quality and realism assessment.

\begin{table}[t]
    \centering
    \caption{Comprehensive evaluation of photorealistic style transfer methods. We report quality metrics (LPIPS, PSNR, SSIM, H-Corr) and inference time. For LUT-based methods, inference time is shown as LUT Generation + LUT Application. Best values are in \textbf{bold}, and second-best are \underline{underlined}.}
    \renewcommand{\arraystretch}{1.2}
    \setlength{\tabcolsep}{4pt}
    \resizebox{\linewidth}{!}{%
    \begin{tabular}{lccccc}
        \toprule
        Method & LPIPS $\downarrow$ & PSNR $\uparrow$ & SSIM $\uparrow$ & H-Corr $\uparrow$ & Inference Time (s) \\
        \midrule
        AdaIN \cite{huang2017adain} & 0.53 & 18.21 & 0.62 & 0.39 & 0.0499 \\
        NLUT \cite{chen2023nlut} & 0.36 & 20.59 & 0.80 & 0.33 & 16.1112 + 0.0003 \\
        Deep Preset \cite{Ho_2021_WACV} & 0.32 & \underline{23.42} & 0.84 & 0.41 & \textbf{0.0002} \\
        ModFlow \cite{larchenko2024color} & 0.28 & 20.13 & 0.85 & 0.33 & \underline{0.0800} \\
        WCT$^2$ \cite{yoo2019photorealistic} & 0.27 & 19.86 & 0.81 & 0.31 & 0.6600 \\
        Neural Preset \cite{NeuralPreset} & \underline{0.19} & 23.03 & \underline{0.89} & \underline{0.44} & N/A \\
        \textbf{SA-LUT (Ours)} & \textbf{0.12} & \textbf{25.29} & \textbf{0.92} & \textbf{0.51} & 0.2128 + 0.0100 \\
        \bottomrule
    \end{tabular}%
    }
    \label{tab:quantitative_and_speed}
\end{table}

Table~\ref{tab:quantitative_and_speed} presents a comprehensive evaluation of our method against state-of-the-art photorealistic style transfer approaches on both quality metrics and inference speed. Our SA-LUT substantially outperforms all baseline methods across all quality metrics. Notably, we achieved a $66.7\%$ reduction in LPIPS compared to NLUT~\cite{chen2023nlut}, a previous LUT-based method. Our approach also shows impressive performance in structural preservation (SSIM: 0.92) and fidelity (PSNR: 25.29). The strong performance in histogram correlation (H-Corr) indicates that our method successfully captures the color distribution of the target style while maintaining content structure.

In terms of computational efficiency, SA-LUT exhibits a balanced speed-quality trade-off. The LUT generation time (0.2128s) is significantly faster than NLUT (16.1112s), representing a 75$\times$ speedup for the style pre-processing stage. This acceleration is particularly important for applications where users frequently change style references. While some methods like Deep Preset offer faster raw inference, they lack the adaptivity and quality of our approach as demonstrated by our quality metrics.

It is worth noting that once the 4D LUT is generated for a particular style, it can be repeatedly applied to different content images with minimal computational cost, since we only need to recompute the context generator. This separation of style encoding and application makes our approach particularly suitable for real-time applications where a fixed style needs to be applied to multiple images or video frames.

\begin{figure*}[h!]
  \centering
  \includegraphics[width=\textwidth]{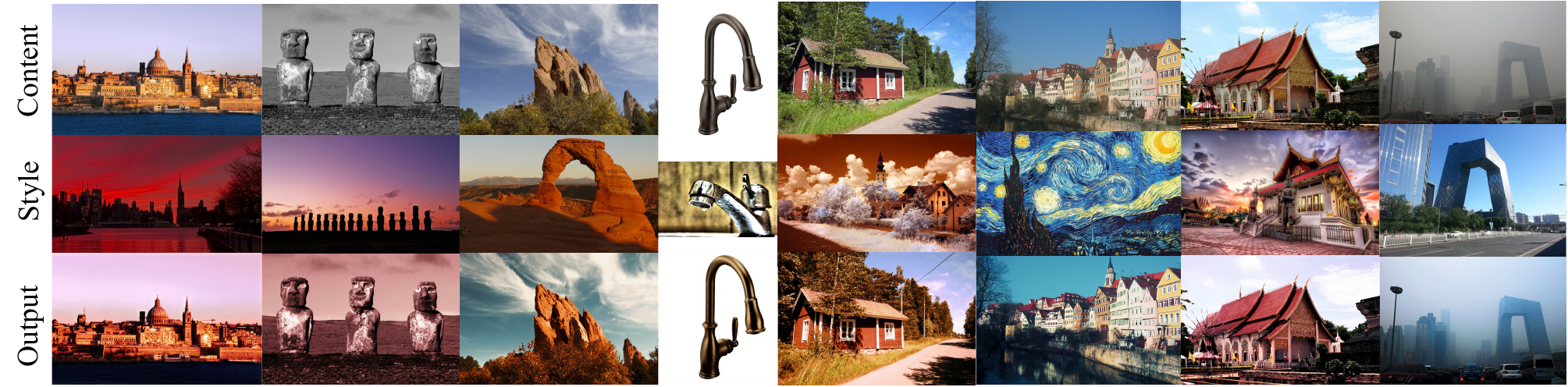}
  \caption{Qualitative results of SA-LUT on image pairs from the DPST dataset, showcasing generalization to diverse styles and lighting conditions.}
  \label{fig:salut_dpst}
\end{figure*}

\subsubsection{Qualitative Results}
As illustrated in Figure \ref{fig:visualize1}, our SA-LUT method effectively transfers the color mood of the style image while preserving the fine structural details of the content image. Unlike some baseline models \cite{NeuralPreset, Ho_2021_WACV}, which, despite producing natural-looking results, may fail to fully capture the style’s color distribution (e.g., the subdued colors in row 2), SA-LUT achieves a more faithful color transfer. Furthermore, our method demonstrates superior local adaptation, particularly in challenging scenarios involving complex textures and scenes, such as trees, leaves, and skies (see row 4), where our context-aware approach enables region-specific grading. Notably, SA-LUT avoids common artifacts, including color blocking and structural distortions, ensuring high-quality stylized outputs. To assess the generalization capabilities of SA-LUT beyond our PST50 benchmark, we also evaluated its performance on the DPST dataset. As shown in Fig~\ref{fig:salut_dpst} SA-LUT successfully transfers diverse styles while preserving content structure and demonstrating robustness to varied local lighting conditions

\subsubsection{User Studies}
\label{sec:userstudy}
We conducted a user study comparing our method against Neural Preset~\cite{NeuralPreset} (highest performer in quantitative metrics) and NLUT~\cite{chen2023nlut} (the only other LUT-based method). Using 20 randomly selected image pairs from PST50, 133 participants with diverse backgrounds evaluated which result better achieved photorealistic style transfer based on realism and style similarity. As shown in Table~\ref{tab:user_study}, our method was preferred 48.79\% of the time, significantly outperforming Neural Preset (33.60\%) and NLUT (17.61\%), confirming the superior visual quality of our approach.

\begin{table}[t]
\centering
\caption{User study results comparing our method with Neural Preset~\cite{NeuralPreset} and NLUT~\cite{chen2023nlut} across 20 image pairs from the PST50 dataset. Values represent percentage of user preference.}
\begin{tabular}{lccc}
\toprule
\textbf{Method} & \textbf{Avg Preference (\%)} & \textbf{Win Count} \\
\midrule
NLUT & 17.61 & 0/20 \\
Neural Preset & 33.60  & 6/20 \\
Ours & \textbf{48.79}  & \textbf{14}/20 \\
\bottomrule
\end{tabular}

\label{tab:user_study}
\end{table}

\subsection{Ablation Studies}
\label{sec:ablation}

To validate our design choices in SA-LUT, we performed an ablation study on the PST50 dataset using LPIPS ($\downarrow$) and H-Corr($\uparrow$) as perceptual metrics, evaluating: the Context Generator with cross-attention, the number of basis LUTs, and our training strategy.

\subsubsection{Effect of Context Generator and Cross-Attention}
\label{sec:ablation_context}

The Context Generator, with cross-attention, facilitates spatially adaptive style transfer. We evaluated SA-LUT against two variants: (1) \textit{w/o Context Generator} (standard 3D LUT), and (2) \textit{w/o Cross-Attention} (content-derived context map \cite{liu20234d}).

\begin{table}[t]
    \centering
    \caption{Impact of Context Generator and Cross-Attention on perceptual quality (LPIPS $\downarrow$) and histogram correlation (H-Corr $\uparrow$).}
    \begin{tabular}{l c c}
        \toprule
        \textbf{Model Variant} & \textbf{LPIPS $\downarrow$} & \textbf{H-Corr $\uparrow$} \\
        \midrule

        w/o Context Generator & 0.14 & 0.38 \\
        w/o Cross-Attention   & 0.13 & 0.46 \\
        SA-LUT            & \textbf{0.12} & \textbf{0.51} \\
        \bottomrule
    \end{tabular}

    \label{tab:ablation_context}
\end{table}

\begin{figure}[ht]
  \centering
  \includegraphics[width=\columnwidth]{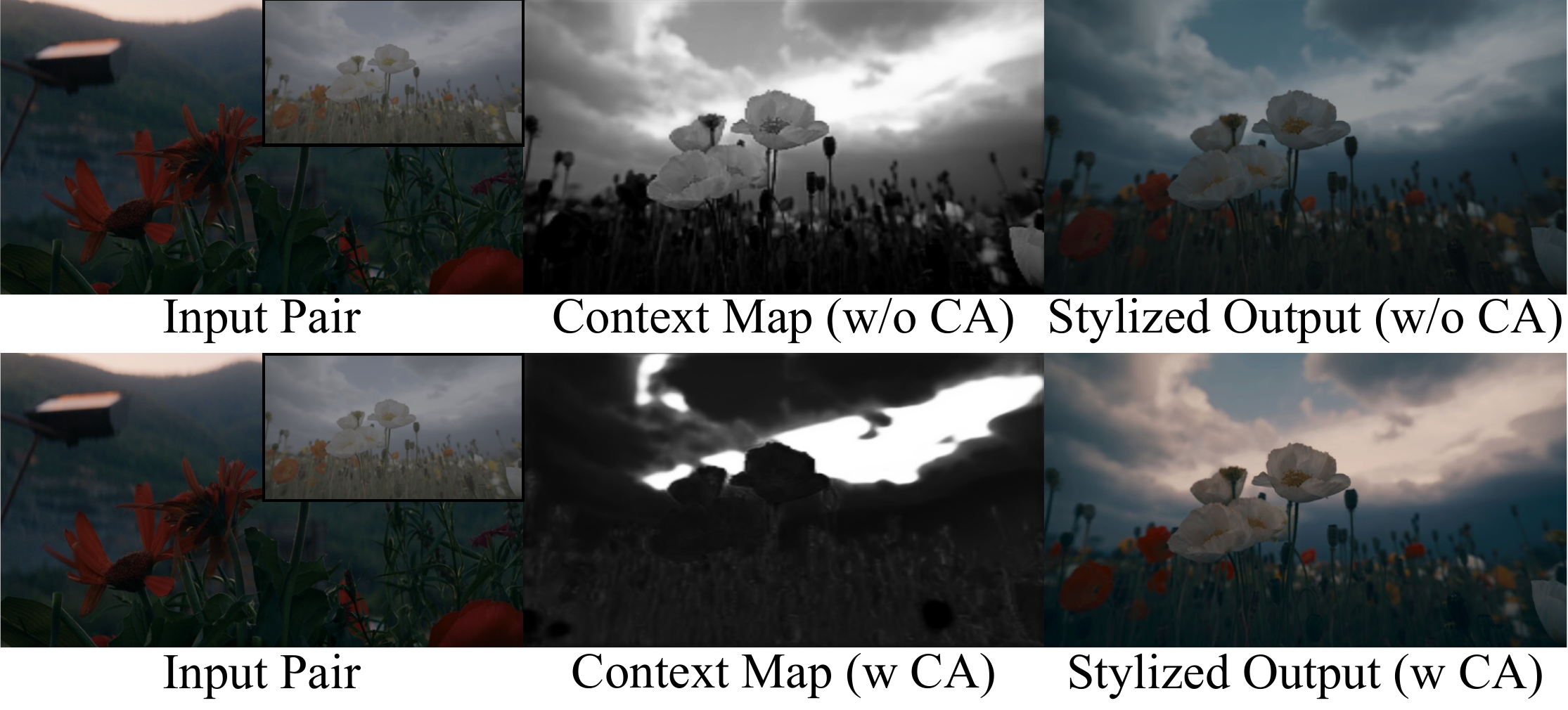}
  \caption{Visual comparison of our methods w/ and w/o cross-attention (CA) module. \textbf{Top}: w/o cross-attention – context map lacks regional information, causing suboptimal color transfer. \textbf{Bottom}: w/ cross-attention – captures region-specific illumination for natural results}
  \label{fig:context_compare}
\end{figure}

Table~\ref{tab:ablation_context} demonstrates the importance of the Context Generator and cross-attention. Removing the Context Generator increases LPIPS (0.12$\rightarrow$ 0.14) and reduces H-Corr to 0.37, confirming standard 3D LUTs are insufficient for spatially adaptive transformations. Similarly, removing cross-attention reduced performance (LPIPS: 0.13 and H-Corr: 0.46), highlighting the necessity of content-style feature interactions for accurate style mapping. Figure~\ref{fig:context_compare} visually confirms this: models without cross-attention fail to capture nuanced illumination variations, while SA-LUT enables region-specific color transformations, preserving structural details like the flowers in Figure~\ref{fig:context_compare}.

\subsubsection{Number of Basis LUTs}
\label{sec:ablation_luts}
\begin{table}[t]
\centering
\caption{Effect of the number of basis LUTs on performance metrics.}
\begin{tabular}{c|c|c}
\toprule
\textbf{Number of Basis LUTs} & \textbf{LPIPS $\downarrow$} & \textbf{H-Corr $\uparrow$} \\
\midrule
32  & 0.14 & 0.41 \\
64  & \textbf{0.12} & \textbf{0.51} \\
128 & 0.13 & 0.47 \\
256 & 0.13 & 0.39 \\
\bottomrule
\end{tabular}

\label{tab:ablation_luts}
\end{table}

Table \ref{tab:ablation_luts} shows performance reaches optimal metrics at 64 LUTs (LPIPS of 0.12 and H-Corr of 0.51). Beyond this point, performance plateaus or slightly degrades, suggesting overfitting or redundancy.
Fewer LUTs (32) lead to significantly worse metrics, indicating insufficient model capacity. These results justify our choice of 64 basis LUTs, balancing expressivity, quality, and efficiency.

\subsubsection{Effect of Training Strategy and Adversarial Loss}
\label{sec:ablation_unpaired}
To evaluate our training strategy, we compared our model with two baseline methods training with only synthetic and real style data, respectively. As shown in Table \ref{tab:ablation_training}, our SA-LUT model achieves the best performance with lowest LPIPS (0.12) and highest H-Corr (0.51). The Synthetic Only variant shows degraded performance, while the Real Style Only variant failed to converge despite TV and monotonicity regularization. These results validate our dual-stream training approach, where synthetic data provides supervision for basic transformations while real style data with adversarial learning captures complex style characteristics.

\begin{table}[t]
\centering
\caption{Impact of training strategy on perceptual quality (LPIPS $\downarrow$) and histogram correlation (H-Corr $\uparrow$)}
\begin{tabular}{l c c}
\toprule
\textbf{Training Variant} & \textbf{LPIPS $\downarrow$} & \textbf{H-Corr $\uparrow$} \\
\midrule
Real Style Only             & N/A & N/A \\
Synthetic Only               & 0.14 & 0.44 \\
SA-LUT     & \textbf{0.12} & \textbf{0.51} \\
\bottomrule
\end{tabular}

\label{tab:ablation_training}
\end{table}

\subsection{Limitations}
\label{sec:limitations}

\begin{figure}[t]
\centering
\includegraphics[width=\columnwidth]{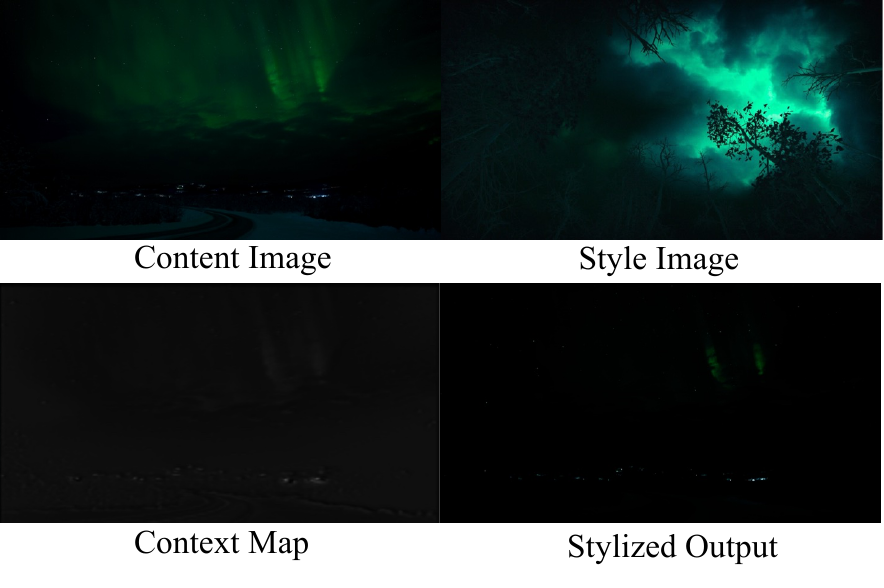}
\caption{Failure case under extreme lighting conditions.}
\label{fig:failure}
\end{figure}

Despite SA-LUT's strong performance, we identify several limitations that suggest directions for future research: (1) Challenging lighting conditions: As shown in Figure~\ref{fig:failure}, our method struggles with severely under/overexposed content images, where the Context Generator cannot produce meaningful spatial maps. Cross-attention fails to establish valid content-style feature correspondences when visual information is insufficient, resulting in poor stylization. (2) Semantic mismatch: Our cross-attention mechanism becomes less effective when content and style images differ dramatically in semantics, resulting in more global, less spatially-adaptive transformations. (3) Temporal consistency: While enabling real-time 4K processing, our method exhibits subtle frame-to-frame variations during rapid scene changes;

\section{Conclusion}
\label{sec:conclusion}

We have presented a novel Spatial Adaptive 4D Look-Up Table (SA-LUT) model for photorealistic style transfer tasks that extends traditional 3D LUTs with a context dimension, enabling spatially adaptive color transformations while maintaining hardware compatibility. Our approach, combining cross-attention between content and style features with a 4D LUT representation, achieves state-of-the-art performance on our PST50 benchmark, reducing perceptual distance by 66.7\% compared to previous LUT-based methods. Beyond style transfer, the principles of our spatially adaptive framework could benefit various image enhancement tasks requiring context-sensitive transformations in computational photography and film post-production.

\noindent\textbf{Social Impacts.} 
Our approach could potentially facilitate image manipulation that might be used for deceptive purposes or misrepresentation. We emphasize that SA-LUT is designed as a complementary tool for creative professionals rather than a replacement, supporting artistic workflows while maintaining the crucial role of human judgment in visual media creation.

 \small \bibliographystyle{ieeenat_fullname} \bibliography{main}

\clearpage
\appendix
\onecolumn

{\Large \bf \centering Supplementary Material \par}

\vspace{2em}

\setcounter{equation}{0}
\setcounter{figure}{0}
\setcounter{table}{0}
\setcounter{section}{0}
\renewcommand{\theequation}{S\arabic{equation}}
\renewcommand{\thefigure}{S\arabic{figure}}
\renewcommand{\thetable}{S\arabic{table}}
\renewcommand{\thesection}{S\Roman{section}}
\renewcommand{\thetable}{\Alph{table}}
\renewcommand{\thefigure}{\Alph{figure}}
\renewcommand{\thesection}{\Alph{section}}

\setcounter{table}{0}
\setcounter{figure}{0}
\setcounter{section}{0}
\begin{figure*}[h]
    \centering
    \includegraphics[width=0.95\textwidth]{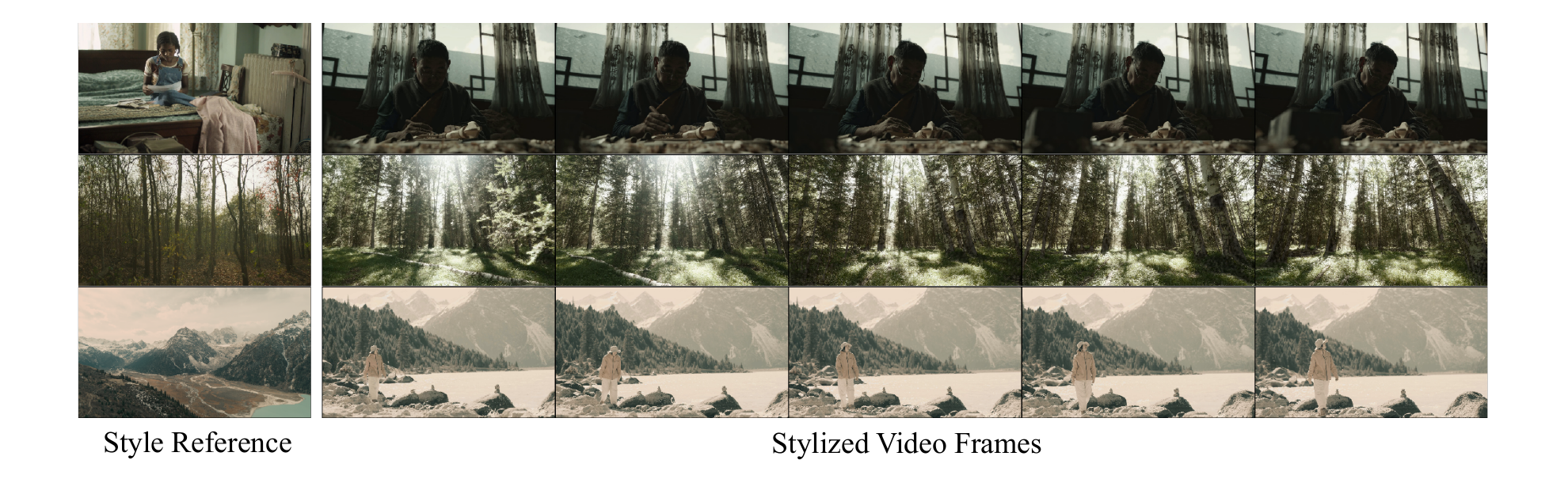}
    \caption{Selected frames from video stylization tests.}
    \label{fig:video_frames}
\end{figure*}

\section{Style2Log Model Details}
\label{style2log}

In this section, we provide detailed information about the Style2Log model used to generate synthetic log-space images from style references, as mentioned in the main paper.

\subsection{Overview}

The Style2Log model is a specialized neural network designed to transform standard images into their log-space representations by learning from style references. This model enables us to leverage unpaired data by generating synthetic training pairs that capture complex color grading characteristics found in professional photography and cinematography.

\subsection{Architecture Components}






To achieve higher-quality outputs, we integrate a NAFNet-based~\cite{chen2022simple} refinement network into Style2Log. Our implementation employs a NAFNet with a width of 32, four middle blocks, and block configurations of [1,1,1,2] for the encoder and [1,1,1,1] for the decoder.

The refinement network receives the initial LUT-transformed image and generates the final log-space representation, enhancing both local details and global consistency.

We trained the Style2Log model using a curated dataset of log images combined with various LUTs. The training objective is formulated as a weighted combination of multiple loss functions:

\begin{equation} \mathcal{L}_\text{total} = \lambda_1 \mathcal{L}_1 + \lambda_2 \mathcal{L}_\text{Perc}, \end{equation}

where $\mathcal{L}_1$ denotes the $\mathcal{L}_1$ loss and $\mathcal{L}_\text{Perc}$ represents the perceptual loss.

\section{Visualization Results for Video Stylization}
\label{sec:video_stylization}

Figure \ref{fig:video_frames} shows selected frames from our video stylization tests. For complete demonstrations of these results, please refer to the attached video.

\section{Visualization Results for PST50 (Paired)}
\label{sec:pst50_paired}
\begin{figure*}[t]
    \centering
    \includegraphics[width=0.92\textwidth]{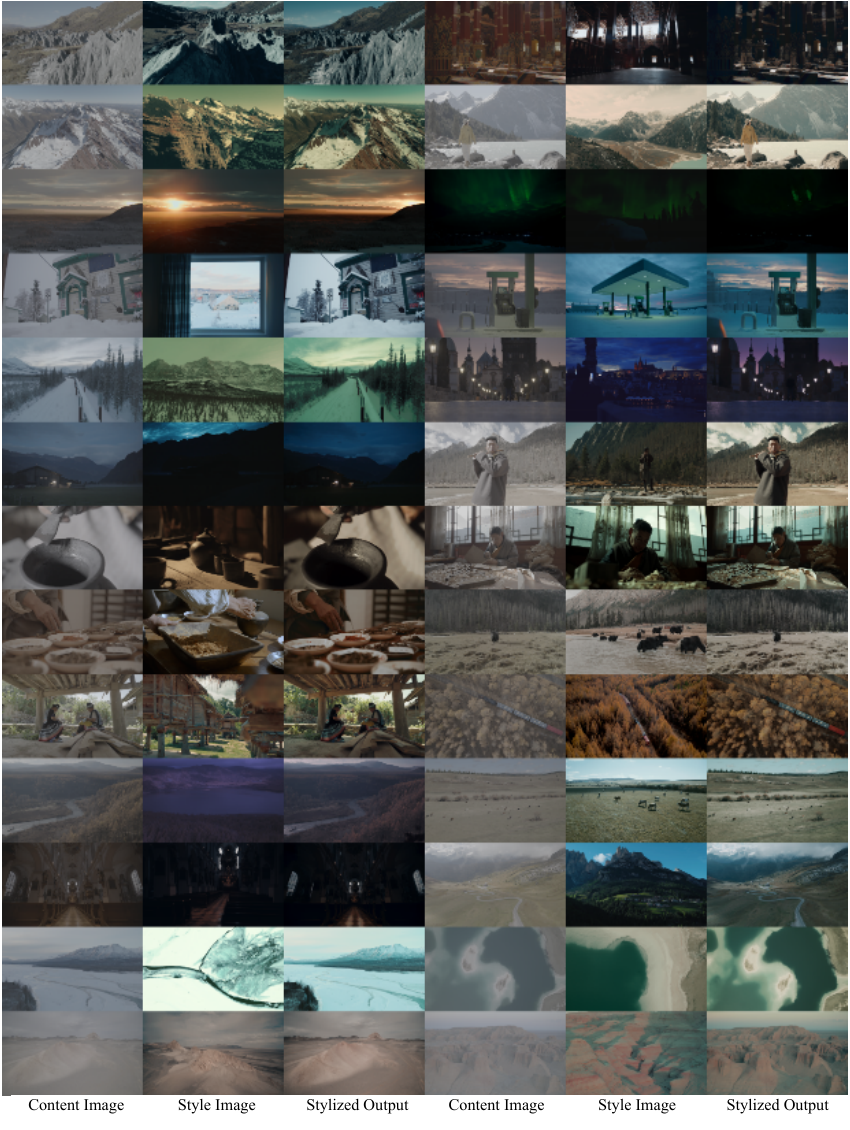}
    \caption{Stylization results on PST50 paired test set (page 1).}
    \label{fig:pst50_paired_1}
\end{figure*}

\begin{figure*}[t]
  \centering
  \includegraphics[width=0.92\textwidth]{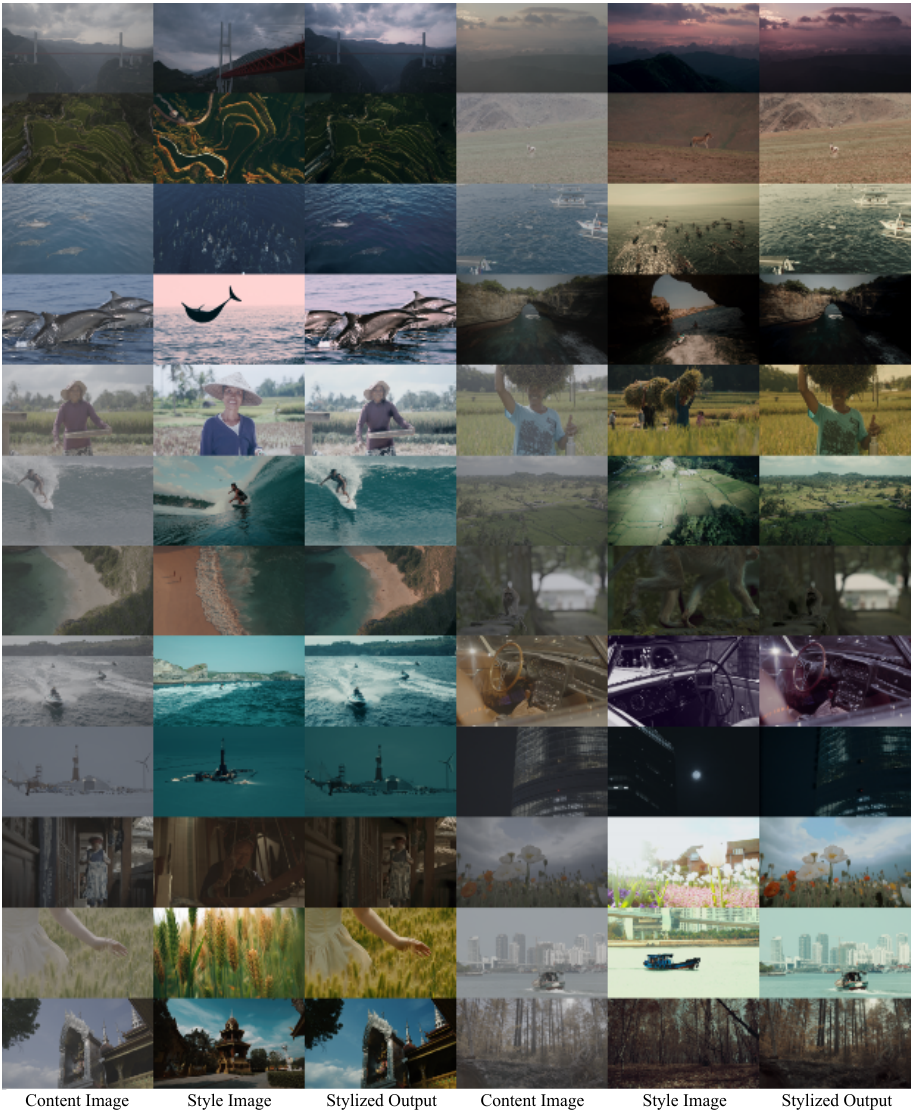}
  \caption{Stylization results on PST50 paired test set (page 2).}
  \label{fig:pst50_paired_2}
\end{figure*}

Figure \ref{fig:pst50_paired_1} and Figure \ref{fig:pst50_paired_2} show additional stylization results on the paired branch of our PST50 benchmark using SA-LUT.
\section{Visualization Results for PST50 (Unpaired)}
\label{sec:pst50_unpaired}
\begin{figure*}[t]
    \centering
    \includegraphics[width=0.92\textwidth]{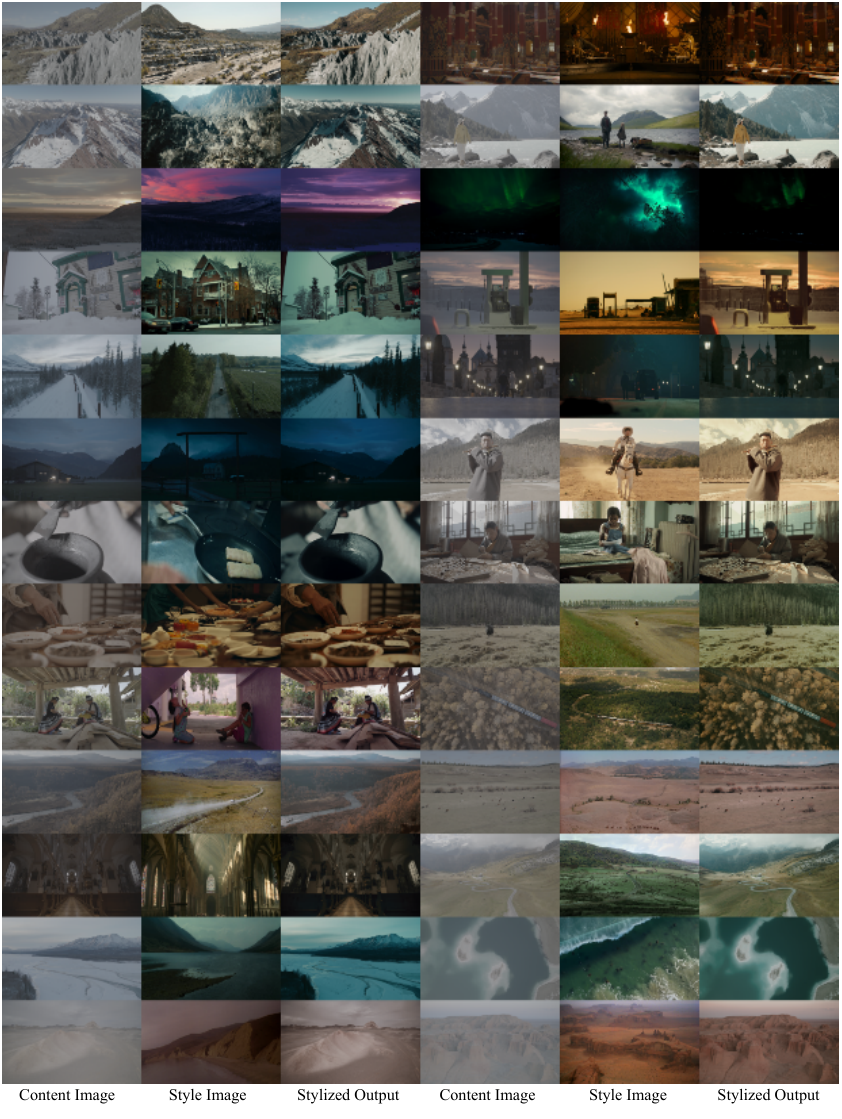}
    \caption{Stylization results on PST50 unpaired test set (page 1).}
    \label{fig:pst50_unpaired_1}
\end{figure*}

\begin{figure*}[t]
    \centering
  \includegraphics[width=0.92\textwidth]{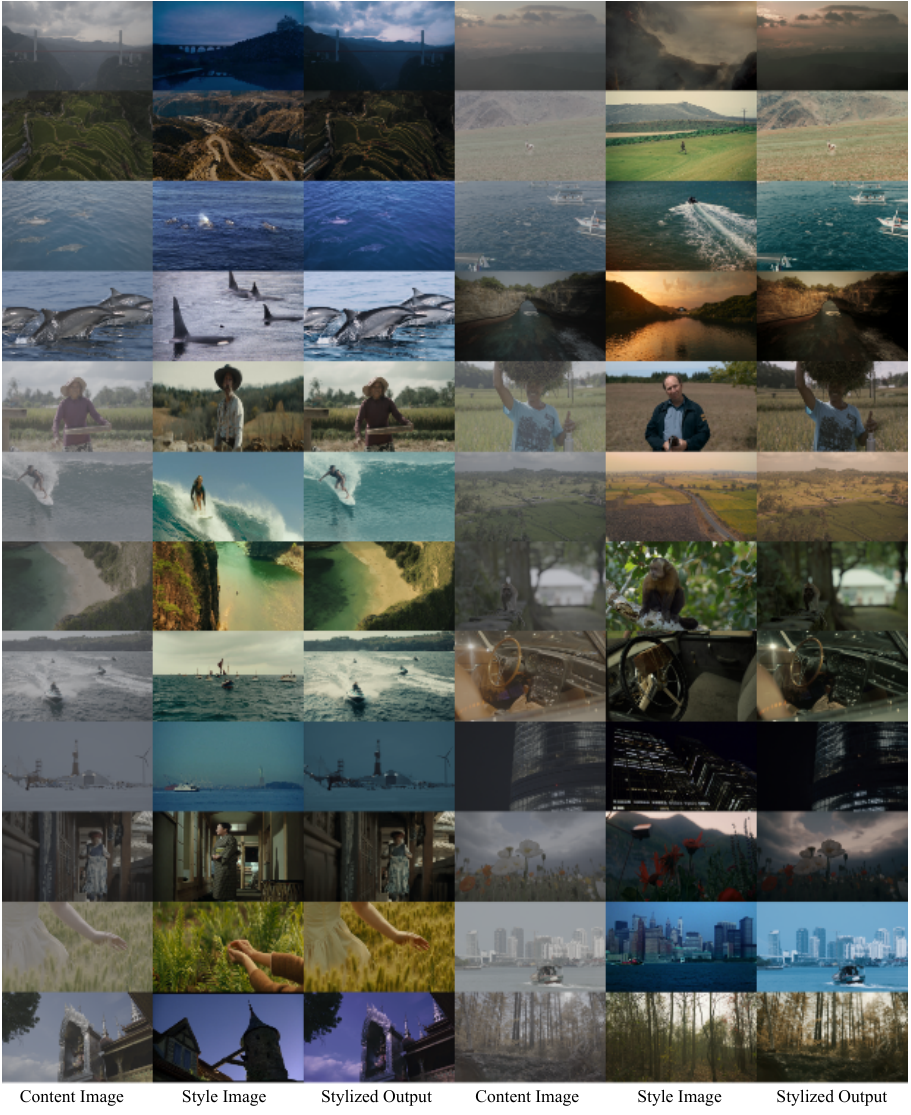}
    \caption{Stylization results on PST50 unpaired test set (page 2).}
    \label{fig:pst50_unpaired_2}
\end{figure*}

Figure \ref{fig:pst50_unpaired_1} and Figure \ref{fig:pst50_unpaired_2} present additional stylization results on the unpaired branch of our PST50 benchmark using SA-LUT.

\end{document}